\documentclass[runningheads,a4paper]{llncs}
\usepackage{amssymb}
\usepackage{amsmath}
\usepackage{graphicx}
\usepackage{url}
\usepackage{adjustbox}
\usepackage{array}
\usepackage[labelformat=empty]{subfig}
\usepackage{color,soul} 
\sethlcolor{white} 
\soulregister\ref{7} 
\soulregister\cite{7} 
\usepackage{float} 
\usepackage[colorlinks]{hyperref}
\usepackage[colorinlistoftodos]{todonotes}
\usepackage{booktabs} 
\usepackage{rotating}
\usepackage{multirow}
\usepackage{transparent}
\usepackage{tablefootnote}
\usepackage[subtle,margins=normal,leading=normal]{savetrees} 
\usepackage{bm} 
\usepackage{colortbl}
\usepackage{tabulary}

\newcommand\mycaptionsize{\small}
\begin{document}
\mainmatter 
\title{A multi-scale pyramid of 3D fully convolutional networks for abdominal multi-organ segmentation}

\titlerunning{A multi-scale pyramid of 3D fully convolutional networks...}

%

\author{Holger R. Roth\inst{1}\thanks{\textit{Contact:} \email{rothhr@mori.m.is.nagoya-u.ac.jp} or \email{kensaku@is.nagoya-u.ac.jp}}
	\and Chen Shen\inst{1} 
	\and Hirohisa Oda\inst{1} 
	\and Takaaki Sugino\inst{1} 
	\and Masahiro Oda\inst{1}
	\and Yuichiro Hayashi\inst{1} 
	\and Kazunari Misawa\inst{2}
	\and Kensaku Mori\inst{1,3,4}
}
\authorrunning{H. R. Roth et al.}

\institute{Graduate School of Informatics, Nagoya University, Nagoya, Japan\\
	\and
	Aichi Cancer Center, Nagoya, Japan\\
	\and
	Information Technology Center, Nagoya University, Nagoya, Japan\\
	\and
	Research Center for Medical Bigdata, National Institute of Informatics, Tokyo, Japan\\
}

\maketitle
\begin{abstract}
Recent advances in deep learning, like 3D fully convolutional networks (FCNs), have improved the state-of-the-art in dense semantic segmentation of medical images. However, most network architectures require severely downsampling or cropping the images to meet the memory limitations of today's GPU cards while still considering enough context in the images for accurate segmentation. In this work, we propose a novel approach that utilizes auto-context to perform semantic segmentation at higher resolutions in a multi-scale pyramid of stacked 3D FCNs. We train and validate our models on a dataset of manually annotated abdominal organs and vessels from 377 clinical CT images used in gastric surgery, and achieve promising results with close to 90\% Dice score on average. For additional evaluation, we perform separate testing on datasets from different sources and achieve competitive results, illustrating the robustness of the model and approach.
\end{abstract}
\section{Introduction}
Multi-organ segmentation has attracted considerable interest over the years. The recent success of deep learning-based classification and segmentation methods has triggered widespread applications of deep learning-based semantic segmentation in medical imaging \cite{milletari2016v,cciccek20163d}. Many methods focused on the segmentation of single organs like the prostate \cite{milletari2016v}, liver \cite{christ2016automatic}, or pancreas \cite{roth2016spatial,zhou2017fixed}. Deep learning-based multi-organ segmentation in abdominal CT has also been approached recently in works like \cite{hu2017automatic,gibson2017towards}. Most of these methods are based on variants of fully convolutional networks (FCNs) \cite{long2015fully} that either employ 2D convolutions on orthogonal cross-sections in a slice-by-slice fashion \cite{christ2016automatic,roth2016spatial,zhou2017fixed,zhou2017deep} or 3D convolutions \cite{milletari2016v,cciccek20163d,gibson2017towards}.
A common feature of these segmentation methods is that they are able to extract features useful for image segmentation directly from the training imaging data, which is crucial for the success of deep learning. This avoids the need for handcrafting features that are suitable for detection of individual organs.

However, most network architectures require severely downsampling or cropping the images for 3D processing to meet the memory limitations of today's GPU cards \cite{milletari2016v,gibson2017towards} while still considering enough context in the images for accurate segmentation of organs. 

In this work, we propose a multi-scale 3D FCN approach that utilizes a scale-space pyramid with auto-context to perform semantic image segmentation at a higher resolution while also considering large contextual information from lower resolution levels. We train our models on a large dataset of manually annotated abdominal organs and vessels from pre-operative clinical computed tomography (CT) images used in gastric surgery and evaluate them on a completely unseen dataset from a different hospital, achieving a promising performance compared to the state-of-the-art. 

\begin{figure}[htb]
	\centering
	\adjincludegraphics[width=0.95\columnwidth]{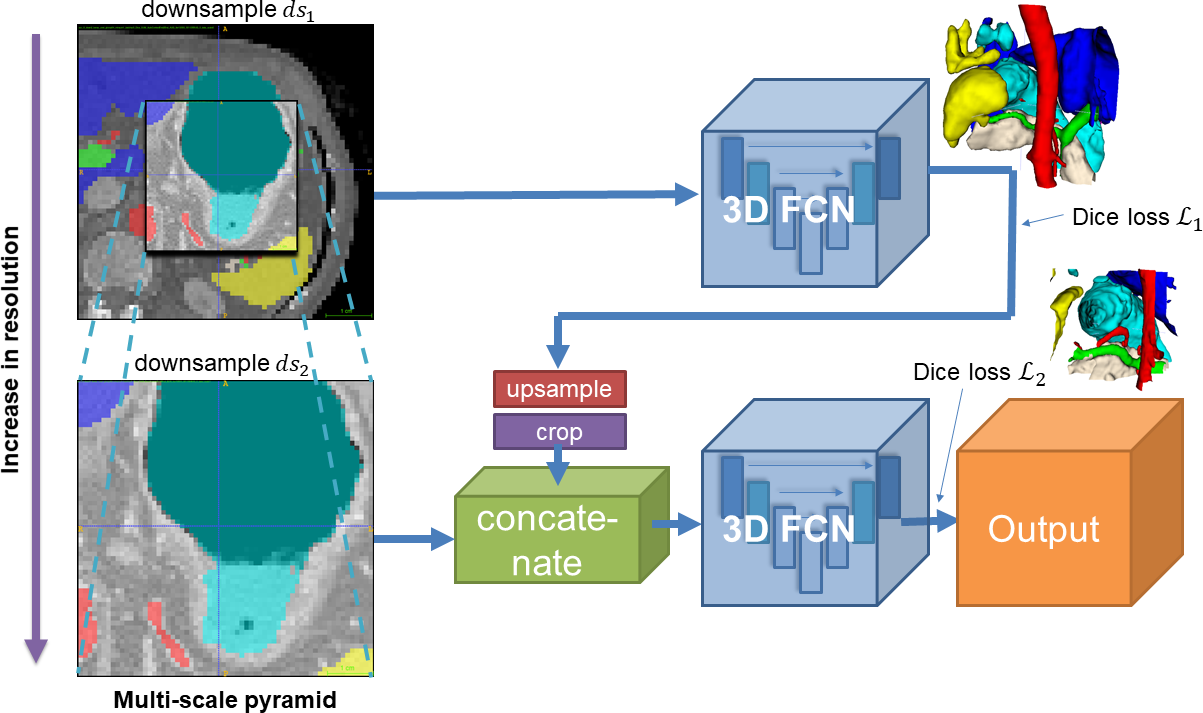}
	\caption{\mycaptionsize Multi-scale pyramid of 3D fully convolutional networks (FCNs) for multi-organ segmentation. The lower-resolution-level 3D FCN predictions are upsampled, cropped and concatenated with the inputs of a higher resolution 3D FCN. The Dice loss is used for optimization at each level and training is performed end-to-end. \label{fig:network}}
\end{figure}

Our approach is shown schematically in Fig. \ref{fig:network}. We are influenced by classical scale-space pyramid \cite{adelson1984pyramid} and auto-context ideas \cite{tu2010auto} for integrating multi-scale and varying context information into our deep learning-based image segmentation method. Instead of having separate FCN pathways for each scale as explored in other work \cite{chen2016attention,salehi2017auto}, we utilize the auto-context principle to fuse and integrate the information from different image scales and different amounts of context. This helps the 3D FCN to integrate the information of different image scales and image contexts at the same time. Our model can be trained end-to-end using modern deep learning frameworks. \hl{This is in contrast to previous work which utilized auto-context using a separately trained models for brain segmentation \cite{salehi2017auto}}.

\hl{In summary,} our contributions are \textbf{1)} introduction of a multi-scale pyramid of 3D FCNs; \textbf{2)} improved segmentation of fine structures at higher resolution; \textbf{3)} end-to-end training of multi-scale pyramid FCNs showing improved performance and good learning properties. We perform a comprehensive evaluation on a large training and validation dataset, plus unseen testing on data from different hospitals and public sources, showing promising generalizability.
\section{Methods}
\subsection{3D fully convolutional networks}
Convolutional neural networks (CNN) have the ability to solve challenging classification tasks in a data-driven manner. Fully convolutional networks (FCNs) are an extension to CNNs that have made it feasible to train models for pixel-wise semantic segmentation in an end-to-end fashion \cite{long2015fully}. In FCNs, feature learning is purely driven by the data and segmentation task at hand and the network architecture. Given a training set of images and labels $\mathbf{S} = \left\{(X_n,L_n),\ n = 1,\dots,N\right\}$, where $X_n$ denotes a CT image and $L_n$ a ground truth label image, the model can train to minimize a loss function $\mathcal{L}$ in order to optimize the FCN model $f(I,\Theta)$, where $\Theta$ denotes the network parameters, including the convolutional kernel weights for hierarchical feature extraction.

While efficient implementations of 3D convolutions and growing GPU memory have made it possible to deploy FCN on 3D biomedical imaging data \cite{milletari2016v,cciccek20163d}, image volumes are in practice often cropped and downsampled in order for the network to access enough context to learn an effective semantic segmentation model while still fitting into memory. Our employed network model is inspired by the fully convolutional type 3D U-Net architecture proposed in {\c{C}}i{\c{c}}ek et al. \cite{cciccek20163d}.  

\paragraph{The 3D U-Net architecture} \hl{is based on U-Net proposed in \cite{ronneberger2015u} and} consists of analysis and synthesis paths with four resolution levels each. It utilizes deconvolution \cite{long2015fully} (also called transposed convolutions) to remap the lower resolution and more abstract feature maps within the network to the denser space of the input images. This operation allows for efficient dense voxel-to-voxel predictions. Each resolution level in the analysis path contains two $3 \times 3 \times 3$ convolutional layers, each followed by rectified linear units (ReLU) and a $2 \times 2 \times 2$ max pooling with strides of two in each dimension. In the synthesis path, the convolutional layers are replaced by deconvolutions of $2 \times 2 \times 2$ with strides of two in each dimension. These are followed by two $3 \times 3 \times 3$ convolutions, each followed by ReLU activations. Furthermore, 3D U-Net employs shortcut (or skip) connections from layers of equal resolution in the analysis path to provide higher-resolution features to the synthesis path \cite{cciccek20163d}. The last layer contains a $1\times 1\times 1$ convolution that reduces the number of output channels to the number of class labels $K$. This architecture has over 19 million learnable parameters and can be trained to minimize the average Dice loss derived from the binary case in \cite{milletari2016v}:
\begin{equation}
\mathcal{L}\left(X,\Theta,L\right)=- \frac{1}{K}\sum_{k=1}^K \left( \frac{2\sum_{i}^{N} p_{i,k}l_{i,k}}{\sum_{i}^{N} p_{i,k}+\sum_{i}^{N} l_{i,k}} \right).
\label{equ:dice_loss}
\end{equation} 
Here, $p_{i,k} \in [0,\dots,1]$ represents the continuous values of the \textit{softmax} 3D prediction maps for each class label $k$ of $K$ and $l_{i,k}$ the corresponding ground truth value in $L$ at each voxel $i$.
\subsection{Multi-scale auto-context pyramid approach}
\begin{figure*}[htb]
	\centering
	\begin{tabular}{cccc}
		\subfloat[g.t. (scale 0)]{\rotatebox{0}{\adjincludegraphics[valign=c,width=0.24\textwidth]{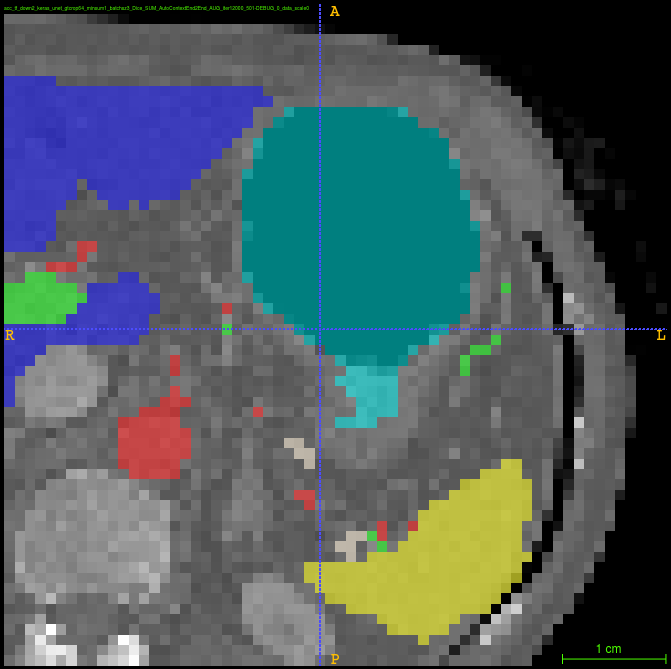}}} &
		\subfloat[pred. (scale 0)]{\rotatebox{0}{\adjincludegraphics[valign=c,width=0.24\textwidth]{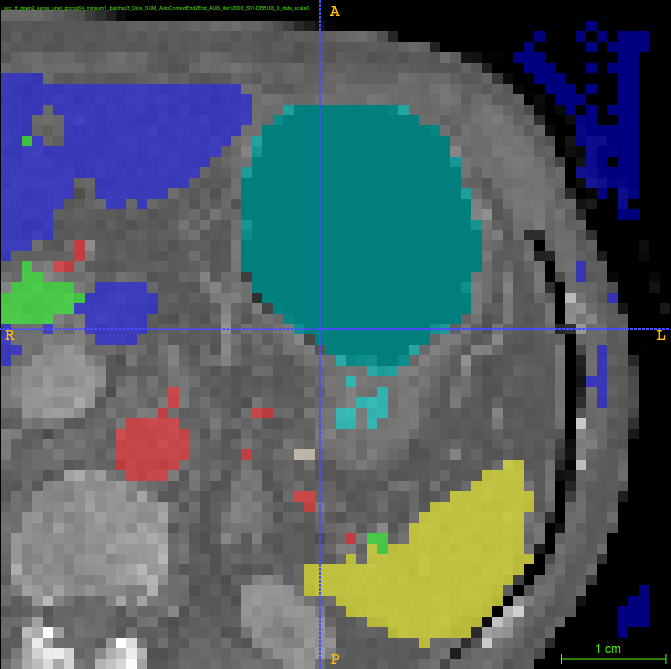}}} &  
		\subfloat[g.t. (scale 1)]{\rotatebox{0}{\adjincludegraphics[valign=c,width=0.24\textwidth]{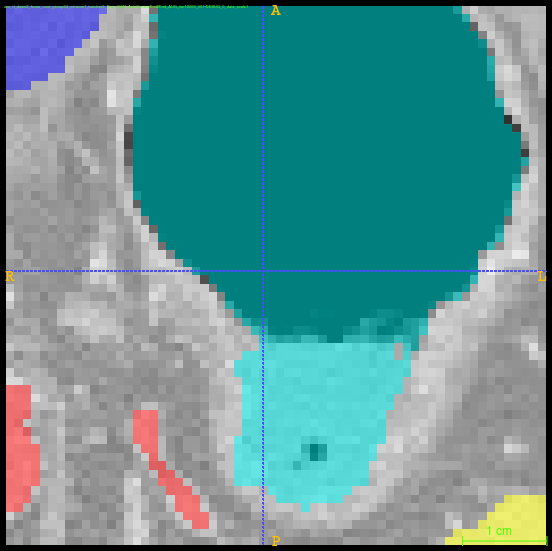}}} &  
		\subfloat[pred. (scale 1)]{\rotatebox{0}{\adjincludegraphics[valign=c,width=0.24\textwidth]{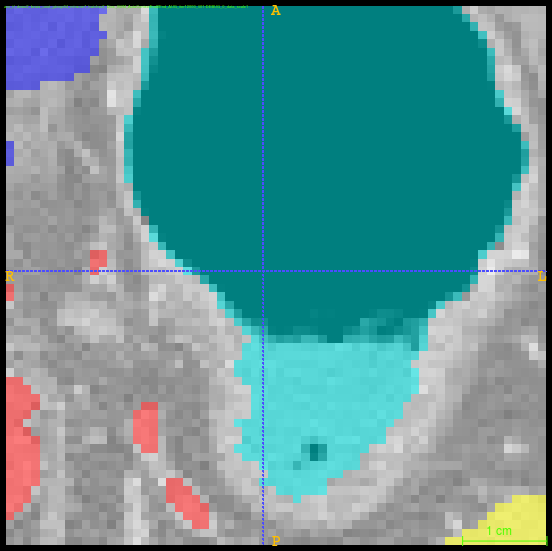}}} \\		
		
		\subfloat[g.t. (scale 0)]{\rotatebox{0}{\adjincludegraphics[valign=c,width=0.24\textwidth]{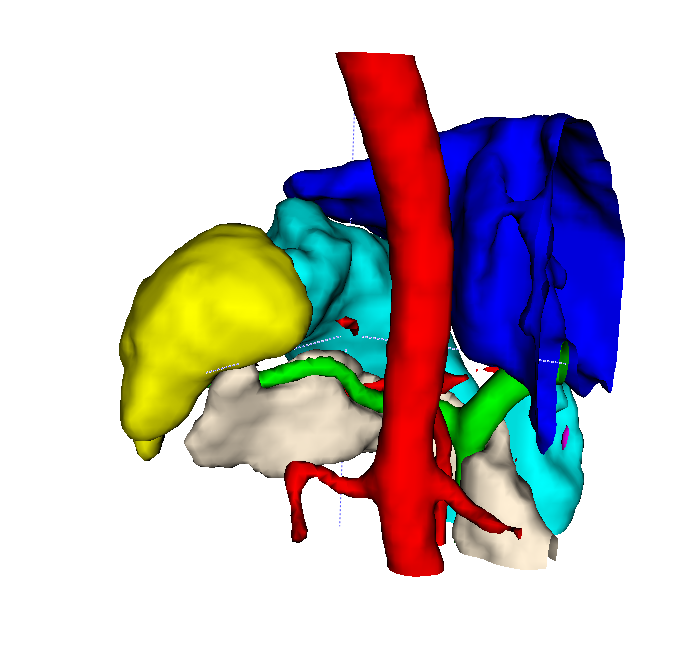}}} &
		\subfloat[pred. (scale 0)]{\rotatebox{0}{\adjincludegraphics[valign=c,width=0.24\textwidth]{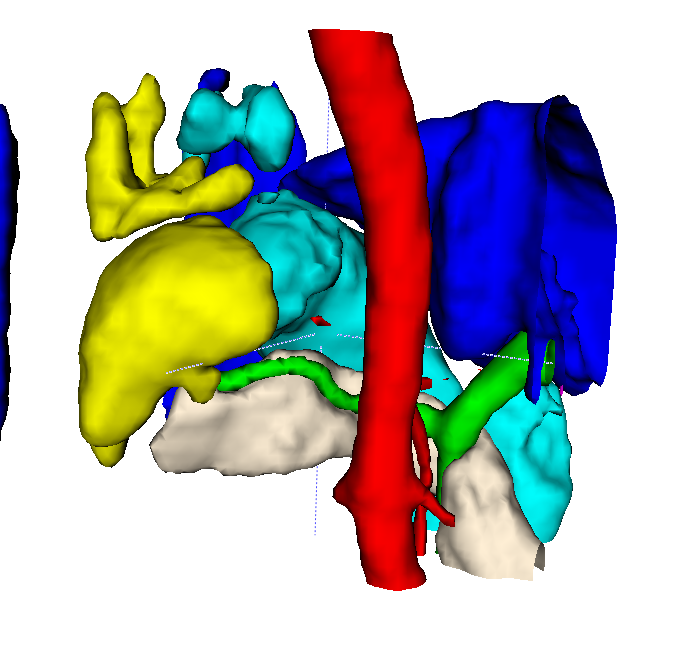}}} &  
		\subfloat[g.t. (scale 1)]{\rotatebox{0}{\adjincludegraphics[valign=c,width=0.24\textwidth]{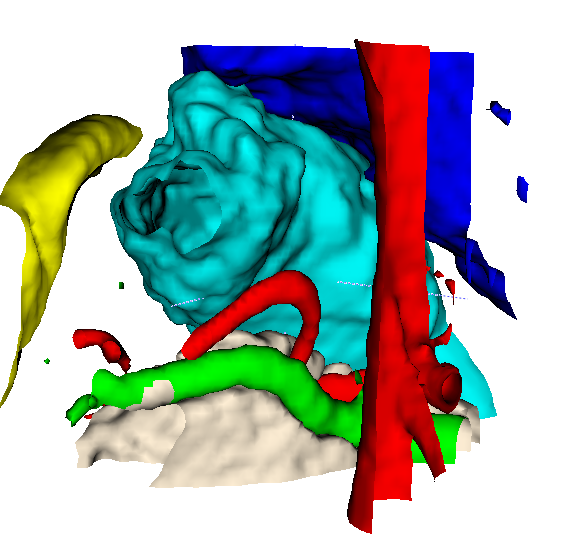}}} &  
		\subfloat[pred. (scale 1)]{\rotatebox{0}{\adjincludegraphics[valign=c,width=0.24\textwidth]{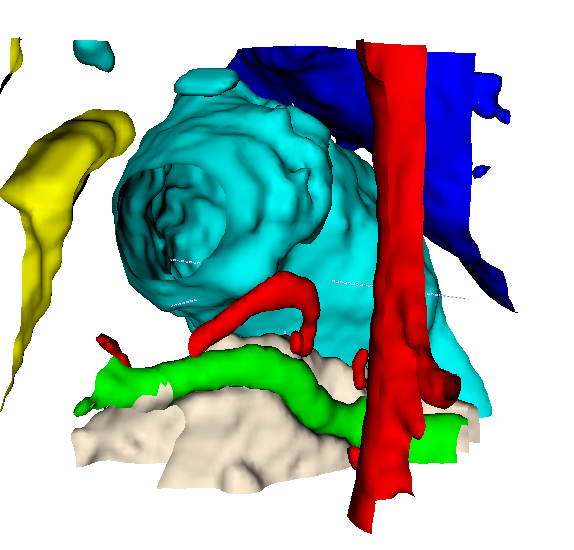}}}
	\end{tabular}
    \begin{center}
    	\includegraphics[width=100mm]{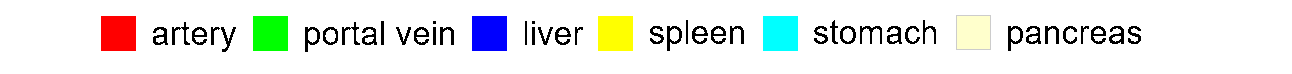}
    \end{center}
	\caption{\mycaptionsize Axial CT images and 3D surface rendering with ground truth (g.t.) and predictions overlaid. We show the two scales used in our experiments. Each scale's input is of size $64\times64\times64$ in this setting. \label{fig:scales}}
\end{figure*}
To effectively process an image at higher resolutions, we propose a method that is inspired by the auto-context algorithm \cite{tu2010auto}. Our method both captures the context information at lower resolution downsampled images and learns more accurate segmentations from higher resolution images in two levels of a scale-space pyramid $\mathbf{F} = \left\{\left(f_s(X_s,\Theta_s)\right),\ s = 1,\dots,S\right\}$, with $S$ being the number of levels $s$ in our multi-scale pyramid, and $X_s$ being one of the multi-scale input subvolumes at each level $s$.  

In the first level, the 3D FCN is trained on images of the lowest resolution in order to capture the largest amount of context, downsampled with a factor of $ds_1=2S$ and optimized using the Dice loss $\mathcal{L}_1$. This can be thought of as a form of deep supervision \cite{lee2015deeply}. In the next level, we use the predicted segmentation maps as a second input channel to the 3D FCN while learning from the images at a higher resolution, downsampled by a factor of $ds_2 = ds_1/2$, and optimized using Dice loss $\mathcal{L}_2$. For input to this second level of the pyramid, the previous level prediction maps are upsampled by a factor of 2 and cropped in order to spatially align with the higher resolution levels. These predictions can then be fed together with the appropriately cropped image data as a second channel. This approach can be learned end-to-end using modern multi-GPU devices and deep learning frameworks with the total loss being 
$
	\mathcal{L}_\mathrm{total} = \sum^L_{s=1}\mathcal{L}_s\left(X_s,\Theta_s,L_s\right)
$. 
This idea is shown schematically in Fig. \ref{fig:network}. The resulting segmentation masks for the two-level case are shown in Fig. \ref{fig:scales}. It can be observed that the second-level auto-context network markedly outperforms the first-level predictions and is able to segment structures with improved detail, especially at the vessels.
\subsection{Implementation \& Training}
We implement our approach in Keras\footnote{\url{https://keras.io/}} using the TensorFlow\footnote{\url{https://www.tensorflow.org/}} backend. The Dice loss [3] is used for optimization with Adam and automatic differentiation for gradient computations. Batch normalization layers are inserted throughout the network, using a mini-batch size of three, sampled from different CT volumes of the training set. We use randomly extracted subvolumes of fixed size during training, such that at least one foreground voxel is at the center of each subvolume. On-the-fly data augmentation is used via random translations, rotations and elastic deformations similar to \cite{cciccek20163d}.
\section{Experiments \& Results}
\begin{figure*}[tb]
	\centering
	\begin{tabular}{ccc}
		\subfloat[Ground truth (axial)]{\rotatebox{0}{\adjincludegraphics[valign=c,width=0.32\textwidth]{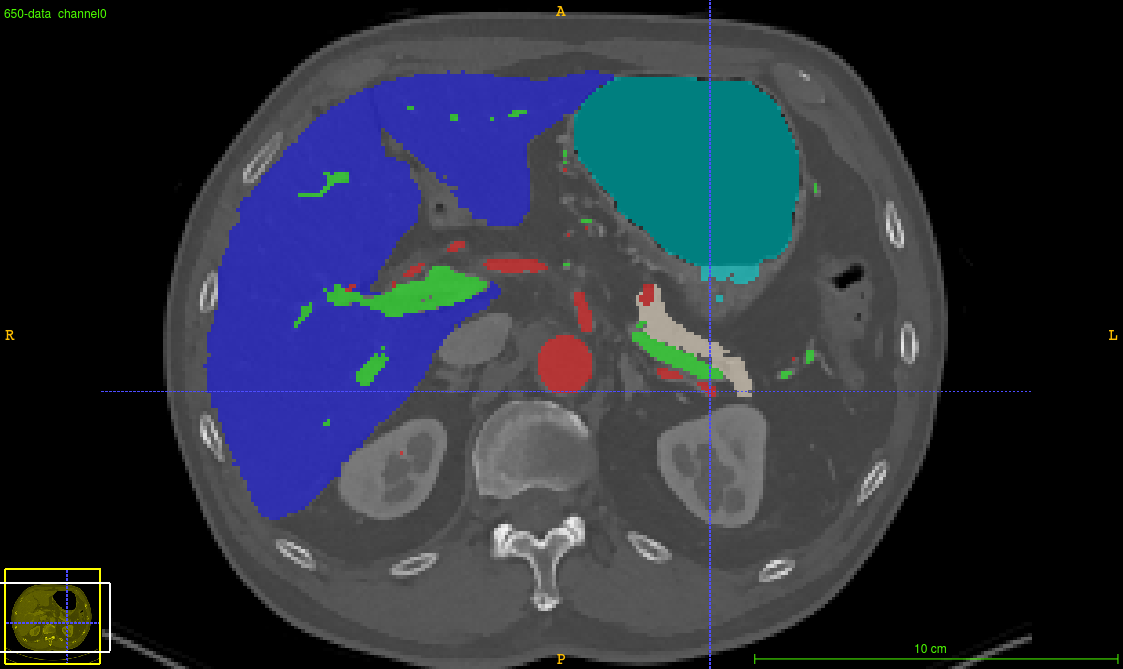}}} &
		\subfloat[first level (upsampled)]{\rotatebox{0}{\adjincludegraphics[valign=c,width=0.32\textwidth]{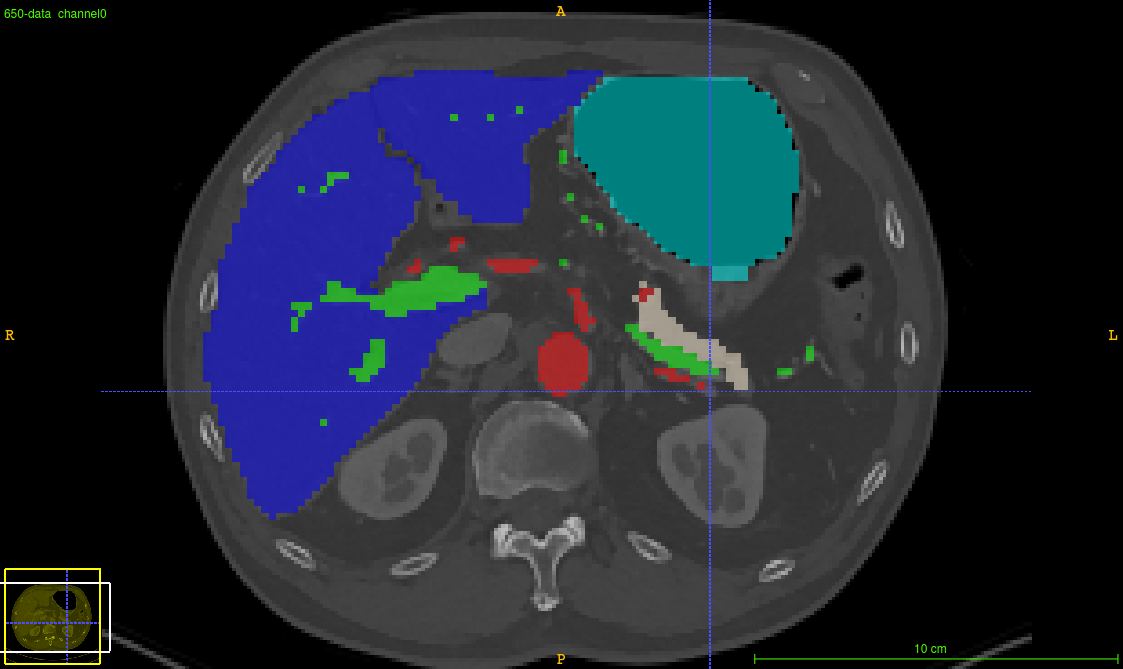}}} &  
		\subfloat[second level (auto-context)]{\rotatebox{0}{\adjincludegraphics[valign=c,width=0.32\textwidth]{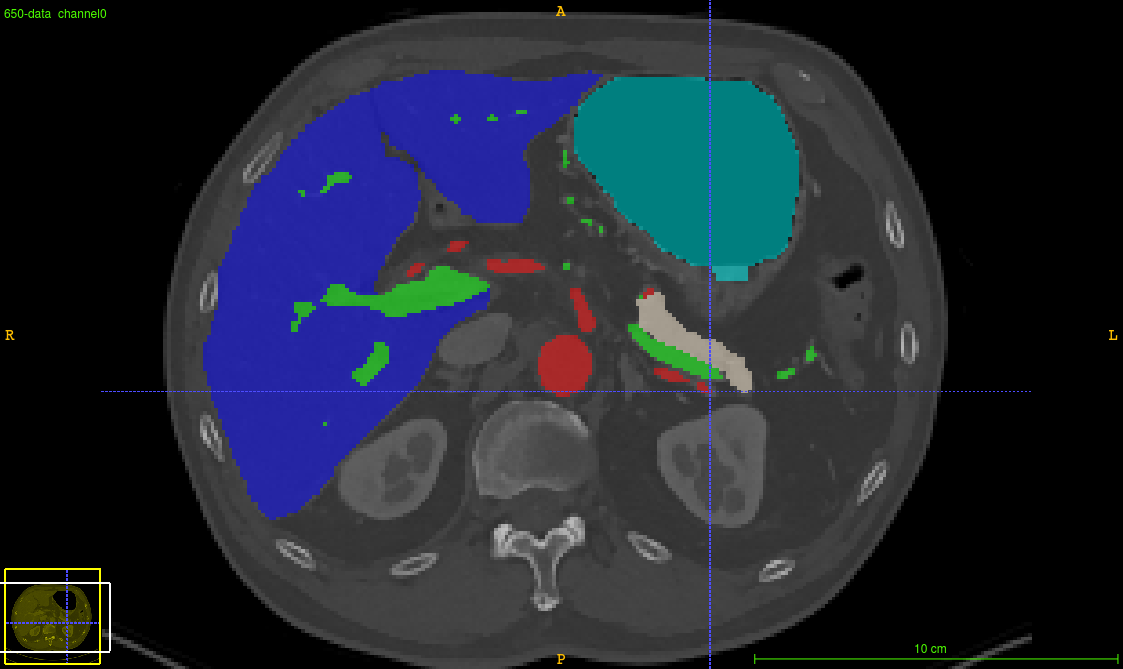}}} \\		
		
		\subfloat[Ground truth (3D)]{\rotatebox{0}{\adjincludegraphics[valign=c,width=0.32\textwidth]{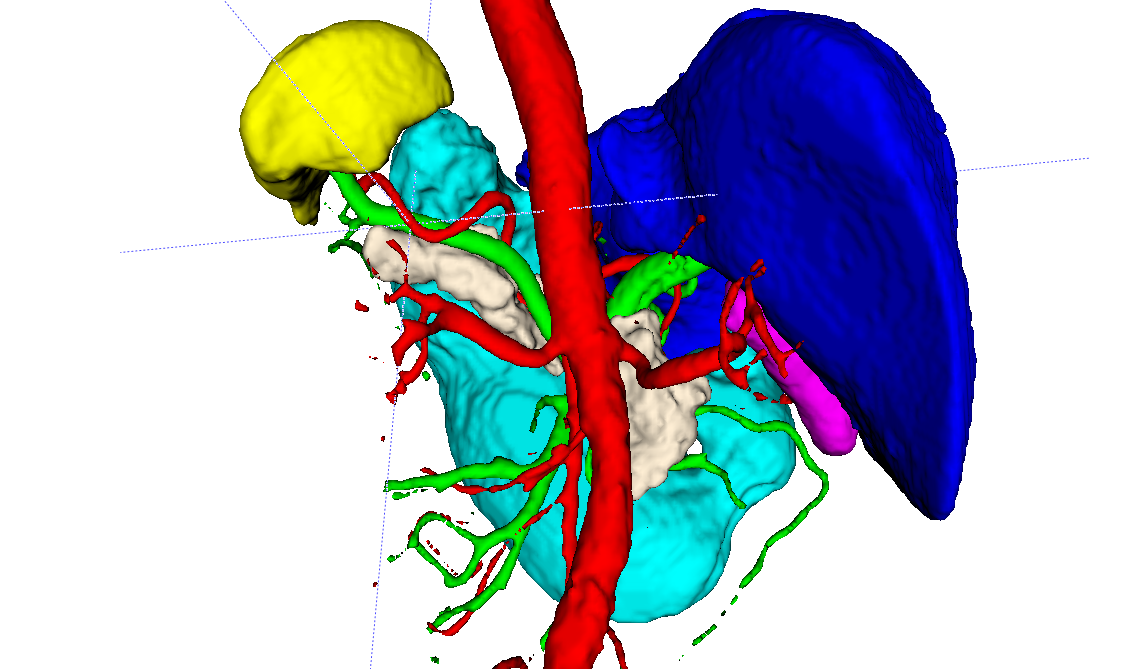}}} &
		\subfloat[first level (upsampled)]{\rotatebox{0}{\adjincludegraphics[valign=c,width=0.32\textwidth]{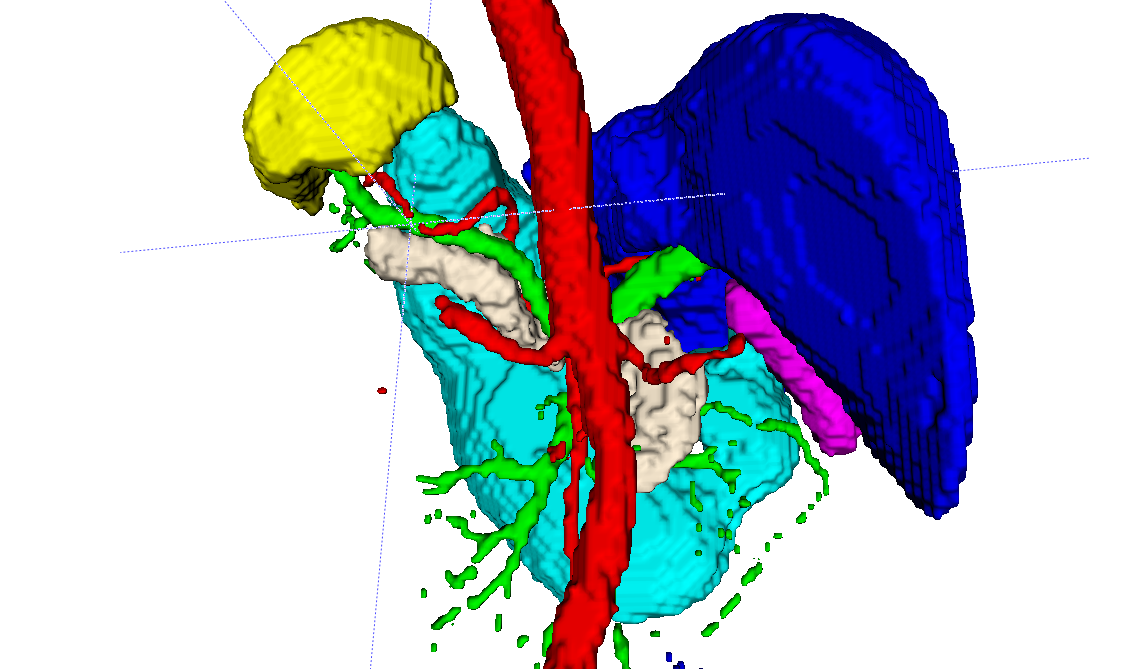}}} &  
		\subfloat[second level (auto-context)]{\rotatebox{0}{\adjincludegraphics[valign=c,width=0.32\textwidth]{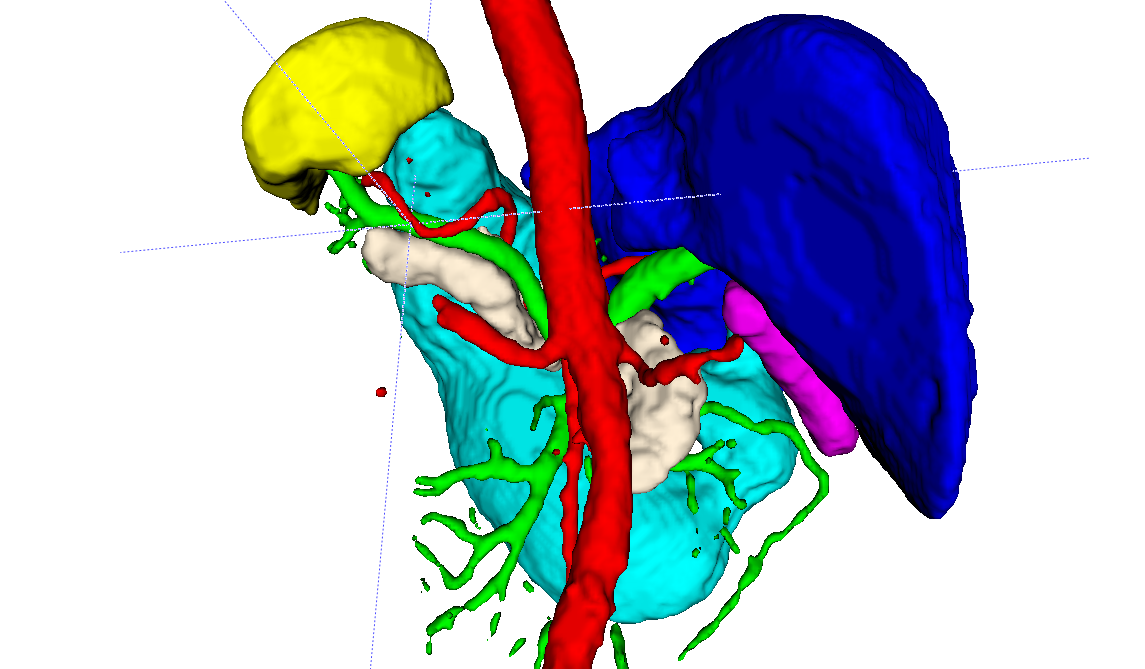}}}
	\end{tabular}
    \begin{center}
    	\includegraphics[width=100mm]{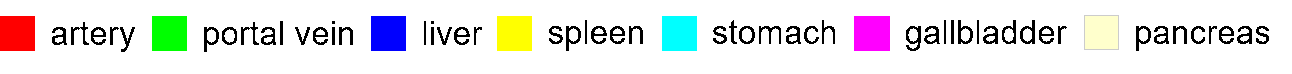}
    \end{center}
	\caption{\mycaptionsize Axial CT images and 3D surface rendering of predictions from two multi-scale levels in comparison with ground truth annotations. In particular, the vessels are segmented more completely and in greater detail in the second level, which utilizes auto-context information in its prediction.\label{fig:results}}
\end{figure*}

In our implementation, a constant input and output size of $64\times64\times64$ randomly cropped subvolumes is used for training in each level. For inference, we employ network reshaping \cite{long2015fully} to more efficiently process the testing image with a larger input size while building up the full image in a tiling approach \cite{cciccek20163d}. The resulting segmentation masks for both levels are shown in Fig. \ref{fig:results}. It can be observed that the second-level auto-context network markedly outperforms the first-level predictions and is able to segment structures with improved detail. All experiments were performed using a DeepLearning BOX (GDEP Advance) with four NVIDIA Quadro P6000s with 24 GB memory each. Training of 20,000 iterations using this unoptimized implementation took several days, while inference on a full CT scan takes just a few minutes on one GPU card.

\paragraph{Data:} Our data set includes 377 contrast-enhanced clinical CT images of the abdomen in the portal-venous phase used for pre-operative planning in gastric surgery. Each CT volume consists of 460--1,177 slices of 512$\times$512 pixels. Voxel dimensions are $[0.59-0.98, 0.59-0.98, 0.5-1.0]$ mm. 
With $S=2$, we downsample each volume by a factor of $ds_1=4$ in the first level and a factor of $ds_2=2$ in the second level. A random 90/10\% split of 340/37 patients is used for training and testing the network. We achieve Dice similarity scores for each organ labeled in the testing cases as summarized in Table \ref{tab:results_val}. \hl{We list the performance for the first level and second level models when utilizing auto-context trained separately or end-to-end, and compare to using no auto-context in the second level. This shows the impact of using or not using the lower resolution auto-context channel at the higher resolution input while training from the same input resolution from scratch.} In our case, each $L_n$ contains $K=8$ labels consisting of the manual annotations of seven anatomical structures (artery, portal vein, liver, spleen, stomach, gallbladder, pancreas), plus background. 


Table \ref{tab:comparison} compares our results to recent literature and also displays the result using an unseen testing dataset from a different hospital consisting of 129 cases from a distinct research study. Furthermore, we test our model on a public data set of 20 contrast-enhanced CT scans.\footnote{We utilize the 20 training cases of the VISCERAL data set (\url{http://www.visceral.eu/benchmarks/anatomy3-open}) as our test set.} 
\begin{table}[htbp]
	\centering
	\caption{\mycaptionsize Comparison of different levels of our model. \hl{End-to-end training gives a statistically significant improvement ($p<0.001$).}}
	\scriptsize	
	\begin{tabular}{lrrrrrrrr}
		\toprule		
		\multicolumn{9}{l}{\textbf{Level 1: Initial (low res)}} \\
		\toprule
		\textbf{Dice} (\%) & \multicolumn{1}{l}{\textbf{artery}} & \multicolumn{1}{l}{\textbf{vein}} & \multicolumn{1}{l}{\textbf{liver}} & \multicolumn{1}{l}{\textbf{spleen}} & \multicolumn{1}{l}{\textbf{stomach}} & \multicolumn{1}{l}{\textbf{gall.}} & \multicolumn{1}{l}{\textbf{pancreas}} & \multicolumn{1}{l}{\textbf{Avg.}} \\
		\midrule
		\textbf{Avg} & 75.4 & 64.0 & 95.4 & 94.0 & 93.7 & 80.2 & 79.8 & 83.2 \\
		\textbf{Std} & 3.9 & 5.4 & 1.0 & 0.8 & 7.6 & 15.5 & 8.5 & 06.1 \\
		\textbf{Min} & 67.4 & 41.3 & 91.5 & 92.6 & 48.4 & 27.3 & 49.7 & 59.7 \\
		\textbf{Max} & 82.3 & 70.9 & 96.4 & 95.8 & 96.5 & 93.5 & 90.6 & 89.4 \\
		\bottomrule
		\multicolumn{9}{l}{\textbf{Level 2: Auto-context}} \\
		\toprule
		\textbf{Dice} (\%) & \multicolumn{1}{l}{\textbf{artery}} & \multicolumn{1}{l}{\textbf{vein}} & \multicolumn{1}{l}{\textbf{liver}} & \multicolumn{1}{l}{\textbf{spleen}} & \multicolumn{1}{l}{\textbf{stomach}} & \multicolumn{1}{l}{\textbf{gall.}} & \multicolumn{1}{l}{\textbf{pancreas}} & \multicolumn{1}{l}{\textbf{Avg.}} \\
		\midrule
		\textbf{Avg} & 82.5 & 76.8 & 96.7 & 96.6 & 95.9 & \textbf{84.4} & 83.4 & 88.1 \\
		\textbf{Std} & 4.1 & 6.4 & 1.0 & 0.7 & 8.0 & 14.0 & 8.4 & 6.1 \\
		\textbf{Min} & 73.3 & 46.3 & 92.9 & 94.4 & 48.1 & 28.0 & 53.9 & 62.4 \\
		\textbf{Max} & 90.0 & 83.5 & 97.9 & 98.0 & 98.7 & 96.0 & 93.4 & 93.9 \\
		\bottomrule
		\multicolumn{9}{l}{\textbf{End-to-End: Auto-context (high-res)}} \\
		\toprule
		\textbf{Dice} (\%) & \multicolumn{1}{l}{\textbf{artery}} & \multicolumn{1}{l}{\textbf{vein}} & \multicolumn{1}{l}{\textbf{liver}} & \multicolumn{1}{l}{\textbf{spleen}} & \multicolumn{1}{l}{\textbf{stomach}} & \multicolumn{1}{l}{\textbf{gall.}} & \multicolumn{1}{l}{\textbf{pancreas}} & \multicolumn{1}{l}{\textbf{Avg.}} \\
		\midrule
		\textbf{Avg} & \textbf{83.0} & \textbf{79.4} & \textbf{96.9} & \textbf{97.2} & \textbf{96.2} & 83.6 & \textbf{86.7} & \textbf{89.0}\\
		\textbf{Std} & 4.4 & 6.7 & 1.0 & 1.0 & 5.9 & 17.1 & 7.4 & 6.2 \\
		\textbf{Min} & 73.2 & 50.2 & 93.5 & 94.9 & 61.4 & 29.7 & 60.0 & 66.1 \\
		\textbf{Max} & 91.0 & 87.7 & 98.3 & 98.7 & 98.7 & 96.4 & 95.2 & 95.1 \\
		\bottomrule
		\multicolumn{9}{l}{\textbf{Level 2: No auto-context (high-res)}} \\
		\toprule
		\textbf{Dice} (\%) & \multicolumn{1}{l}{\textbf{artery}} & \multicolumn{1}{l}{\textbf{vein}} & \multicolumn{1}{l}{\textbf{liver}} & \multicolumn{1}{l}{\textbf{spleen}} & \multicolumn{1}{l}{\textbf{stomach}} & \multicolumn{1}{l}{\textbf{gall.}} & \multicolumn{1}{l}{\textbf{pancreas}} & \multicolumn{1}{l}{\textbf{Avg.}} \\
		\midrule
		\textbf{Avg} & 69.9 &	72.8 &	86.7 &	90.9 &	3.8 &	73.4 &	77.0 &	67.8 \\	
		\textbf{Std} & 6.2 &	7.0 &	6.4 &	5.3 &	1.3 &	22.5 &	10.8 &	8.5 \\
		\textbf{Min} & 59.5 &	47.1 &	69.9 &	75.7 &	0.7 &	7.8 &	36.1 &	42.4 \\
		\textbf{Max} & 82.1 &	82.9 &	95.7 &	97.0 &	7.4 &	95.9 &	90.9 &	78.8 \\
		\bottomrule
		\multicolumn{9}{l}{\footnotesize \hl{*Best average performance is shown in \textbf{bold}.}}
	\end{tabular}%
	\label{tab:results_val}%
\end{table}%

\begin{table}[htbp]
	\begin{minipage}{\textwidth}
	\centering
	\caption{\mycaptionsize \hl{We compare our model trained in an end-to-end fashion to recent work on multi-organ segmentation}. \cite{zhou2017deep} is using a 2D FCN approach with a majority voting scheme, while \cite{gibson2017towards} employs 3D FCN architectures. Furthermore, we list our performance \hl{on an unseen testing dataset from a different hospital and} on the public Visceral dataset without any re-training and compare it to the current challenge leaderboard (LB) best performance for each organ. Note that this table is incomprehensive and direct comparison to the literature is always difficult due to the different datasets and evaluation schemes involved.}
	\scriptsize	
	\begin{tabular}{llrrrrrrrr}
		\toprule
		\textbf{Dice} (\%) & \textbf{train/test} & \textbf{artery} & \textbf{vein} & \multicolumn{1}{l}{\textbf{liver}} & \textbf{spleen} & \multicolumn{1}{l}{\textbf{stomach}} & \textbf{gall.} & \multicolumn{1}{l}{\textbf{pancreas}} & \textbf{Avg.} \\
		\midrule
		\textbf{Ours (end-to-end)} & 340/37 & 83.0 & 79.4 & 96.9 & 97.2 & 96.2 & 83.6 & 86.7 & 89.0 \\
		\hline
		\textbf{Unseen test} & none/129 & -	& - & 95.3 & 93.6 & -	& 80.8	& 75.7 & 86.3 \\
		\midrule		
		\textbf{Gibson et al. \cite{gibson2017towards}} & 72 (8-CV) & -   & -   & 92    & -   & 83    & -   & 66    & 80.3 \\
		\textbf{Zhou et al. \cite{zhou2017deep}}\footnote{Dice score estimated from Intersection over Union (Jaccard index).} & 228/12 & 73.8 & 22.4 & 93.7 & 86.8 &62.4 & 59.6 &56.1	& 65.0 \\
		\textbf{Hu et al. \cite{hu2017automatic}} & 140 (CV) & - & - & 96.0 & 94.2 & - & - & -	& 95.1 \\
		\midrule
		\textbf{Visceral (LB)} & 20/10 & -	& - & 95.0 & 91.1  & -	& 70.6  	& 58.5  & 78.8 \\
		\textbf{Visceral (ours)}\footnote{At the time of writing, the testing evaluation servers of the challenge were not available anymore for submitting results.} & none/20 & -	& - & 94.0 & 87.2  & -	& 68.2	& 61.9  & 77.8 \\
		\bottomrule
	\end{tabular}%
	\label{tab:comparison}%
\end{minipage}
\end{table}%
\section{Discussion \& Conclusion}
The multi-scale auto-context approach presented in this paper provides a simple yet effective method for employing 3D FCNs in medical-imaging settings. No post-processing was applied to any of the network outputs. The improved performance in our approach is effective for all organs tested (apart from the gallbladder, where the differences are not significant). Note that we used different datasets (from different hospitals and scanners) for separate testing. This experiment illustrates our method\rq{}s generalizability and robustness to differences in image quality and populations. Running the algorithms at a quarter to half of the original resolution improved performance and efficiency in this application. While this method could be extended to using a multi-scale pyramid with the original resolution as the final level, we found that the added computational burden did not add significantly to the segmentation performance. The main improvement comes from utilizing a very coarse image (downsampled by a factor of four) in an effective manner. 
In this work, we utilized a 3D U-Net-like model for each level of the image pyramid. However, the proposed auto-context approach should in principle also work well for other 3D CNN/FCN architectures and 2D and 3D image modalities.
 
In conclusion, we showed that an auto-context approach can result in improved semantic segmentation results for 3D FCNs based on the 3D U-Net architecture. While the low-resolution part of the model is able to benefit from a larger context in the input image, the higher resolution auto-context part of the model can segment the image with greater detail, resulting in better overall dense predictions. Training both levels end-to-end resulted in improved performance.
\paragraph{\textbf{Acknowledgments}} \hl{This work was supported by MEXT KAKENHI (26108006, 17H00867, 17K20099) and the JPSP International Bilateral Collaboration Grant.}
\small
\bibliographystyle{splncs} 

\end{document}